# Stress Assessment with Convolutional Neural Network Using PPG Signals

Yasin Hasanpoor
*Advanced Service Robots (ASR) Lab.,
Department of Mechatronics
Engineering,
Faculty of New Sciences and
Technologies,
University of Tehran,*
Tehran, Tehran, Iran
yasin.hasanpoor@ut.ac.ir

Bahram Tarvirdizadeh
*Advanced Service Robots (ASR) Lab.,
Department of Mechatronics
Engineering,
Faculty of New Sciences and
Technologies,
University of Tehran,*
Tehran, Tehran, Iran
bahram@ut.ac.ir

Khalil Alipour
*Advanced Service Robots (ASR) Lab.,
Department of Mechatronics
Engineering,
Faculty of New Sciences and
Technologies,
University of Tehran,*
Tehran, Tehran, Iran
k.alipour@ut.ac.ir

Mohammad Ghamari
*Department of Electrical and Computer
Engineering,
Kettering University,*
Flint, Michigan, USA,
mghamari@kettering.edu

*Abstract*— Stress is one of the main issues of nowadays lifestyle. If it becomes chronic it can have adverse effects on the human body. Thus, the early detection of stress is crucial to prevent its hurting effects on the human body and have a healthier life. Stress can be assessed using physiological signals. To this end, Photoplethysmography (PPG) is one of the most favorable physiological signals for stress assessment. This research is focused on developing a novel technique to assess stressful events using raw PPG signals recorded by Empatica E4 sensor. To achieve this goal, an adaptive convolutional neural network (CNN) combined with Multilayer Perceptron (MLP) has been utilized to realize the detection of stressful events. This research will use a dataset that is publicly available and named wearable stress and effect detection (WESAD). This dataset will be used to simulate the proposed model and to examine the advantages of the proposed developed model. The proposed model in this research will be able to distinguish between normal events and stressful events. This model will be able to detect stressful events with an accuracy of 96.7%.

## I. INTRODUCTION

Although stress can be a reaction to a short-lived situation, and it may even have positive effects in short run, it can be harmful when it becomes chronic. However, chronic stress can have adverse effects. It can disturb the body's natural behavior and as a result of that, it can cause diseases such as heart disorders, obesity, Alzheimer's, diabetes, depression, gastrointestinal problems, and asthma [1]. Chronic stress occurs due to the prolonged repetition of acute stress and it can lead to severe problems for health [2][3]. Stress can be controlled. The management of stress depends on the detection of initial signs of stress early. Early detection of stress can prevent stress-related health problems later in life. Acute stress can be detected by using physiological signals. Among all physiological signals that have been being used recently for stress detection, PPG signal has attracted much attention. PPG signal is collected by a PPG sensor that is embedded into a wearable device [4][5][6]. Therefore, a wearable device incorporating a PPG sensor can be utilized for the detection of acute stress. According to research[7], stressful events can have influences on the PPG signal. Figure 1 shows how stress can impact on the PPG signals.

In recent years, deep neural networks have been utilized for stress detection using PPG signals and have shown more breakthroughs compared to traditional techniques[8]. Initial attempts to develop stress detection techniques using physiological signals were begun in early 2000. In addition, several research works used statistical methods for detection of stress [9][10]. In recent years, many stress classification techniques have been employed for the control and management of stress and these techniques have mainly been used for healthcare purposes. Among classification algorithms, machine learning (ML) approaches like SVM, KNN, random forest, and decision tree networks, take advantage of some of the features of PPG signal to classify stress status [11][12]. In all of the aforementioned works, ML methods have shown 90% and 85% accuracy for 2–classes and 5-classes classification, respectively [13][14][15]. Although, using combined ML methods and Fuzzy Theory were resulted in better performances[16]. In addition, Deep Neural Network (DNN) approaches, have been able to increase the accuracy of stress classification techniques by 6% on average using raw PPG signals [15]. It must be noted that authors in [17] used other physiological signals such as galvanic skin response, human body temperature, body movement, acceleration, and Electromyography in addition to the PPG signal and their results showed an accuracy of above 99%. Nevertheless, the purpose of this study is merely to use PPG signals to detect stress. In addition to stress detection, using deep neural networks has several more applications such as Heart Rate Variability (HRV) monitoring [18], hypertension detection [19], and many more. Previous studies were focused on improving network accuracies for 2-class and 3-class classifications by means of ML networks or DNN methods that were customized for the understudying dataset [10][11][20].

In this paper, an adaptive deep learning method named CNN-MLP has been used to classify five different modes of stress (i.e. 5 modes shown in figure 2). Figure 2 shows all five different classes of stress in a window, where individual subject was selected randomly. More specifically, in this research, an adaptive 1D CNN has been implemented for the stress assessment of one of the subjects selected from the WESAD dataset. The simulated results showed an accuracy of 96.7% to distinguish whether the subject is under stress or does not have any stress. Again, the same subject was evaluated to investigate different levels of stress. The simulated results showed an accuracy of 80.6% when five different states (i.e., baseline, stress, amusement, meditation, and transient) were investigated. Then, the CNN was reconfigured, and as a result of this reconfiguration, the accuracy of the results was increased by 2% on average for preprocessed PPG data. In addition, two different subjects with different baselines were studied to investigate the effects of baseline on CNN accuracy. The simulated results showed 4% reduction in accuracies with changing baselines. Finally, all 15 subjects were investigated. The obtained results showed an accuracy of 92.1% for distinguishing whether all 15 subjects were under the stress or did not go under any stress and again all 15 subjects were evaluated by applying various stress levels; simulated results showed an accuracy of 80.3%. It must be noted that, in this research, the effect of subjects' baseline has been discussed which has not been studied in the previous works. The reason this has been investigated here is due to the fact that the amplitude of PPG signal of a person in normal state may be near to the amplitude of the PPG signal of another person in stress state. This paper has utilized the adaptive CNN-MLP method for PPG signal classification which has been initially introduced by authors in [21] and it was later reconfigured for 1D in [22] for ECG classification. We used this structure for classification of PPG signals for stress detection.

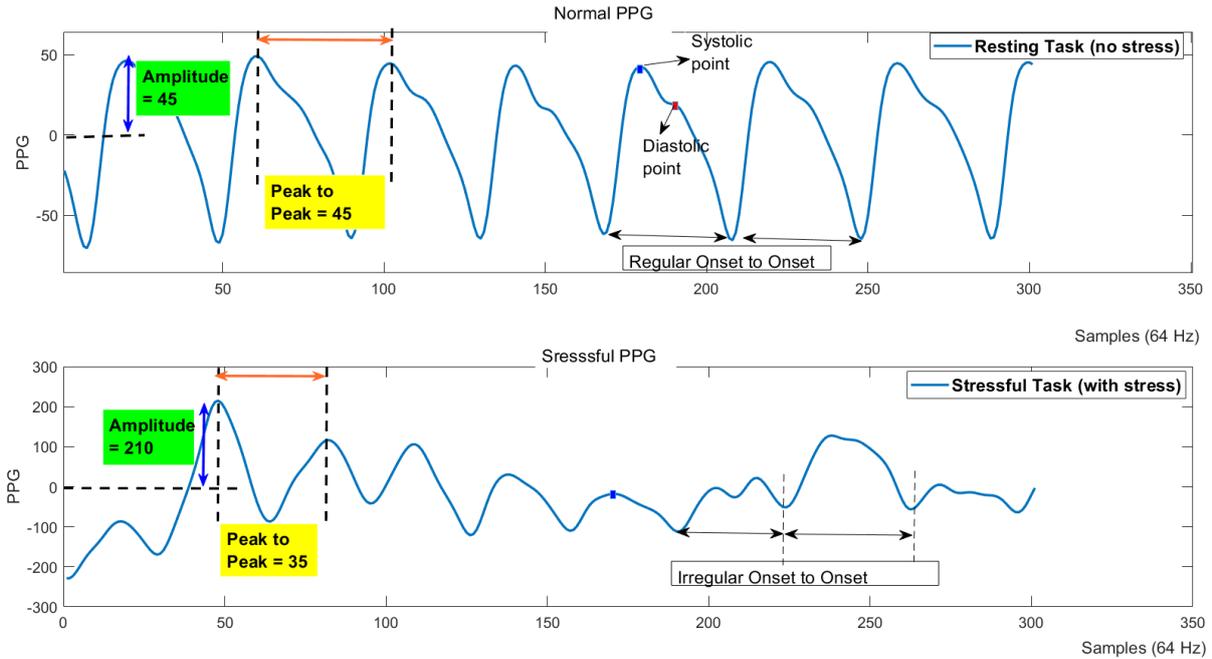

*Figure 1. impact of stress on PPG signals; Normal PPG has a rhythmic structure with regular Onset to Onset, equal Peak to Peak length and distinguishable Systolic and Diastolic points, while stress disrupt these properties and change amplitudes, Peak to Peak, Onset to Onset distances. and other specifications. Besides, Systolic and Diastolic points are not distinguishable in some beats.*

The organization of this paper is structured as follows: in section II, the understudying dataset will be described. Section III will illustrate the dataset preparation and input to the CNN. The design of the adaptive 2D CNN and applied changes to reach adaptive 1D CNN will be discussed in section IV. Section V is on the network implementation for the dataset and parameters of the CNN-MLP and then the baseline effects and accuracies of 2, 3, and 5 classifications will be illustrated.

In section VI and VII the results will be discussed and conclusion will be remarked, respectively.

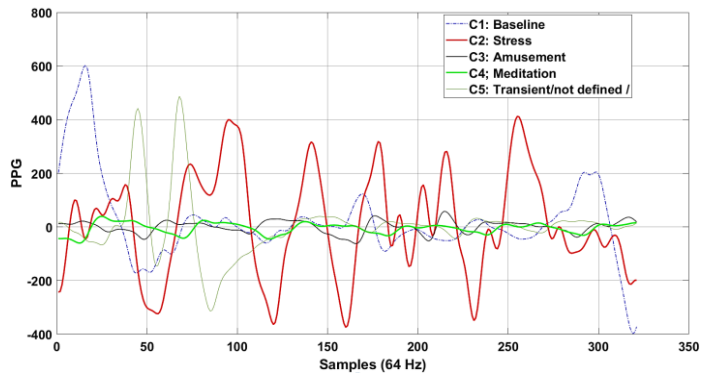

*Figure 2. The effect of stress on PPG signal: The PPG signal has been investigated under five different modes associated with stress during a window of 5-seconds; As can be seen, in stress mode, PPG signal has more irregular properties (e.g., PPG signal has higher amplitude), while these irregularities have not been appeared in meditation mode.*

## II. MATERIALS AND DATA

Wearable stress and affect detection (WESAD) is a publicly available dataset released on 2018 by UCI university [5]. After screening volunteers in terms of pregnancy, being smoker, mental health, …. Motion and physiological data of 17 subjects were recorded from the chest and wrist-worn devices with self-reports during 5 conditions (baseline, amusement, transient, stress, and meditations). It's needed to note that data of subjects 1 and 12 were discarded due to sensor malfunction[13]. Subjects are 12 men and the rest are women. The subjects' average age is 27 ± 4. Recording devices are RespiBAN and Empatica E4. PPG signals that are used in this study were recorded by the E4 wristband with a recording frequency of 64Hz. The steps of the experiment are 5 states done in 2 different protocols as shown in figure 3. The grade of stress is shown by the grade of red and also blue for relax modes. The amusement and stress conditions are aimed at exciting the subjects. During the amusement condition that lasted 392 seconds, the subjects watched a set of eleven funny video clips. In stress conditions, the subjects were exposed to the well-studied Trier Social Stress Test (TSST) [23]. Both stress and amusement conditions were followed by a guided meditation to de-excite the subjects based on controlled breathing exercise.

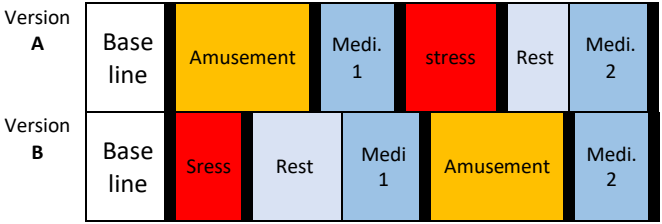

Figure 3. Steps of study in two different versions. The black boxes refer to self-reports filling

Labels are divided into 5 modes: 0 = not defined / transient, 1 = baseline, 2 = stress, 3 = amusement, 4 = meditation. The frequency of labels is 700Hz, whereas the frequency of PPG signals is 64 Hz. PPG signals were captured for 5 different modes and shown in figure 2. It must be noted that, the ground truth was obtained by collecting 5 self-reports of each participant at specified times as shown in figure 3.

## III. PRE-PROCESSING

Data pre-processing is the first step we start with. We will see how preprocessing results in classification accuracy improvement. As mentioned in section I, PPG signals and labels have been sorted by different frequencies. Therefore, labels were synched with PPG signals using downsampling and defining windows frames as shown in figure 4. After that, PPG signal was normalized between -1 and +1. After normalization, moving average and Chebyshev II filters were designed to reduce noises. Noises are mainly due to motion artifacts (MA), sensors' noises, and noises of the environment. Although, there is different unbinding methods for MA noise cancellation, but these methods have not been used in this study because the adaptive CNN can Implicitly discriminate MA noises from clean PPG. Between different filters introduced in [24] and [25], it's observed that the Chebyshev II filter improves the PPG signal quality more effectively than other filters due to its superior frequency selectivity and flat passband [26]. Therefore, the Chebyshev II filter was implemented in preprocessing after the moving average filter that causes signals to be more smooth.

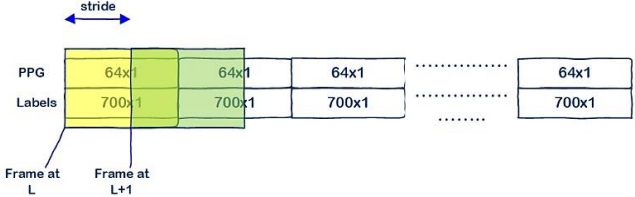

Figure 4. how windows frame moves and compares PPG and corresponding labels

## IV. ADAPTIVE CNN-MLP, STRUCTURE AND DESIGN

As mentioned earlier, the adaptive CNN is used for classification of raw PPG data. Although non-deep methods need PPG feature extraction for classification, the adaptive CNN-MLP can extract features by itself. In this section, we shall make a brief reference to 2D CNN and then we present the design of 1D CNN-MLP changed from 2D adaptive CNN and its backpropagation (BP) training.

### A. 2-D Adaprive CNN

What we mean by the adaptive CNN is that have the freedom of any input layer dimension independent from the CNN parameters and neurons of the CNN in hidden layers are extended such that they are capable of both convolution and downsampling as shown in figure 5. Besides, this structure is such that can be CNN-only without the MLP layers. In figure 5, Kernels (weights) can have different sizes, although they are Kx = Ky = 3 in the figure. The final output of kth neuron in Lth layer is Sk which is subsampled of yk. During the forward propagation (FP), the input signal of each layer has been obtained from the previous layer's neurons as formulated in equation 1.

$$x_k^l = b_k^l + \sum_{i=1}^{N_{l-1}} conv2D(w_{ik}^{l-1} \cdot s_i^{l-1}) \qquad (1)$$

where $x_k^l$ and $b_k^l$ are respectively input and bias of the kth neuron in layer $l$, and $s$ is the output at specified layer and neuron. $w_{ik}^{l-1}$ is the weigh from ith neuron at the specified layer (i.e. $l-1$) to kth neuron at the next layer ($l$). During the BP, three more elements are stored in each neuron those shown in blue blocks in figure 5: Δk (delta error), Δsk (downsampled delta error), and F'(Xk) (derivative of the intermediate output).

We aim that the number of hidden CNN layers can be set to any number. the implementation shown in figure 5 makes this ability possible because the subsampling factor of the last CNN layer (just before the first MLP layer) is set to the dimensions of its input signal automatically. For more details of the structure of 2D adaptive CNN and computations about BP, Weight (kernel), and Bias sensitivity, we refer the audience to reference [22] and [27]. But here we just recapitulate basics that are needed to understand section B.

As the process of BP, the input delta at each layer can be obtained from the next layers and after performing the first BP. As a rule of BP, the delta of each neuron, $\Delta_k^l$, will be formed by the impact of all connected next layer neuron's delta as shown by equations 2

$$\frac{\partial E}{\partial s_k^l} = \Delta s_k^l = \sum_{i=1}^{N_{l+1}} \Delta_i^{l+1} \frac{\partial x_i^{l+1}}{\partial s_k^l} \qquad (2-1)$$

$$x_i^{l+1} = \cdots + s_k^l * w_{ki}^l + \cdots \qquad (2-2)$$

Where the operator * is the conv2D(…) operator without zero padding.

### B. Changes for the Adaptive 1-D CNN Implementation

The main difference is that the convolution equation for 2D (eqn. 1), converts to 1D convolution (eqn. 3). Besides, the kernel sizes and subsampling rates becomes single scalars. The structure of MLP layers are the same for both 1D and 2D CNN and thereby no changes in BP and FP for those layers.

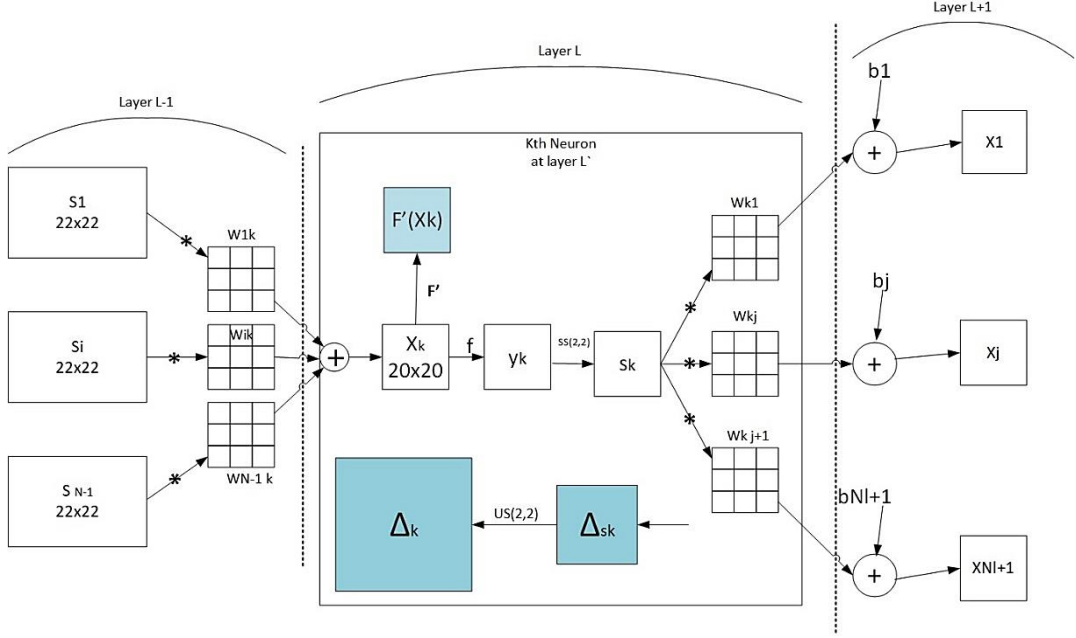

*Figure 5. general structure of Adaptive CNN; blue blocks are parameters that be saved in each neuron*

$$x_k^l = b_k^l + \sum_{i=1}^{N_{l-1}} conv1D(w_{ik}^{l-1} . s_i^{l-1}) \quad (3)$$

The delta error of output Sk becomes 1D computations as eqn.4

$$\Delta S_k^l = \sum_{i=1}^{N_{l-1}} conv1Dz(\Delta_i^{l+1} . rev(w_{ki}^l)) \quad (4)$$

Consequently, the weight and bias sensitivities become equations 5:

$$\frac{\partial E}{\partial w_{ki}^l} = conv1D(S_k^l . \Delta_i^{l+1}) \quad (5-1)$$

$$\frac{\partial E}{\partial b_k^l} = \sum_n \Delta_k^l(n) \quad (5-2)$$

## V. MODEL IMPLEMENTATION AND EXPERIMENTAL SETUP

The goal of this paper is to classify stress by 2 methods:

1- The CNN will be trained by only one person's PPG and therefore we want to classify each person's stress separately. for this purpose, we consider 40% of each person's PPG for training and the remaining for test.

2- we want to classify each person's stress through the PPG of other people. For this purpose, we concatenated all the vectors of each subject's PPG for each class as you can see in table 1 and fed into the CNN.

After Data normalization, to be ensured that the model will not be "stuck" with too many bad batches, shuffling of data after each epoch was considered. In terms of the structures, For the most accurate network, we purposefully used the CNN-MLP in all experiments by N CNN layers, M MLP layers, st stride, FS frame size, fs filter size, and subsampling value of ss. values of these parameters and their results are shown in table 2. the numbers of filters are decided to be 8 and 5 in all CNN and MLP layers respectively since more filters cause no accuracy improvement, but also longer time of runs.

For all experiments, the maximum number of BP iterations is set to be 200, and another stopping criterion is the minimum train classification error level that is set to 1% to prevent overfitting. The model was run on MATLAB.

*Table 1. input vectors of the CNN for 5-classes classification. It's the same with 2 and 3 classifications.*

|  | Subject 2 | Subject 3 | .......... | Subject 17 |
|---|---|---|---|---|
| Class 1 (baseline) | S2_PPG[73152,1] | S3_PPG[72896,1] | .......... | S17_PPG[75520,1] |
| Class 2 (stress) | S2_PPG[39296,1] | S3_PPG[40896,1] | .......... | S17_PPG[46208,1] |
| Class 3 (amusement) | S2_PPG[23104,1] | S3_PPG[23936,1] | .......... | S17_PPG[237440,1] |
| Class 4 (meditation) | S2_PPG[49024,1] | S3_PPG[49792,1] | .......... | S17_PPG[46656,1] |
| Class 5 (transient) | S2_PPG[195456,1] | S3_PPG[213952,1] | .......... | S17_PPG[174784,1] |

To see the effect of baseline, subjects 8 and 15 were selected to check how the difference of baseline changes the accuracy of the classification. Because subject 8 had a stressful day and felt cold in the experiment room. On the other hand, subject 15 didn't really believe the cover story of the stress condition (TSST). This mindset difference, caused the network to have lower accuracy, as shown in Table 2. This result shows that we face low accuracy for stress classification of people with different baselines. Also, mindsets have profound effects on baselines.

## VI. RESULTS

Regarding our purposes as described in section V, the network was first trained to classify the stress of one person (i.e. subject 2. Then the network was implemented for subjects 8 and 15 in order to baseline effects study and finally, classifications were implemented for all 15 subjects. The best results are 94.6% for one subject's 2-class classification while preprocessing caused better accuracy of 96.7% while other research that has worked on stress detection using PPG signals reported 92.1% accuracy[15]. In case of studying 15 subjects together, results are 92.1% and 80.3% respectively for 2-classes and 5 classes classifications. The results of different settings and structures of network can be found in table 2. The 4th training (all subjects with filtration) encountered low accuracies for 3-class classifications (less than 80%), so it is not listed in the table.

*Table 2. summarized results and optimized structures of the CNN training*

| Training Description | No. of classes | No of CNN layers (N) | No. of MLP layers (M) | Frame size (F) | Filter size (f) | Subsampling rate (SS) | Stride (st) | Training accuracy | Testing accuracy | row |
|---|---|---|---|---|---|---|---|---|---|---|
| One subject stress classification - no filtration | 2 | 2 | 3 | 128 | 32 | 2 | 4 | 95.7% | 94.6% | 1 |
| | 3 | 3 | 2 | 64 | | 2 | 4 | 83.0% | 81.4% | 2 |
| | 5 | 3 | 3 | 64 | | 2 | 4 | 82.1% | 80.0% | 3 |
| One subject stress classification - filtered with Chebyshev II and moving average | 2 | 3 | 3 | 64 | 16 | 2 | 4 | 97.0% | 96.7% | 4 |
| | 3 | 2 | 3 | 64 | 16 | 2 | 4 | 84.2% | 82.1% | 5 |
| | 5 | 3 | 3 | 64 | 16 | 2 | 4 | 82.9% | 80.6% | 6 |
| Subjects 8 and 15 stress classification- no filtration | 2 | 2 | 3 | 128 | 32 | 2 | 4 | 91.7% | 90.1% | 7 |
| | 3 | 3 | 3 | 64 | 16 | 2 | 4 | 80.2% | 79.8% | 8 |
| | 5 | 3 | 3 | 64 | 16 | 2 | 4 | 80.0% | 79.7% | 9 |
| All subjects stress classification – with filtration | 2 | 2 | 3 | 128 | 32 | 2 | 4 | 93.7% | 92.1% | 10 |
| | 5 | 3 | 3 | 64 | 16 | 2 | 4 | 80.8% | 80.3 | 11 |

## VII. CONCLUSIONS

The main aim of this work was stress classifications in various classes using the raw PPG signals of the publicly available WESAD dataset. Although the 1D adaptive CNN that was implemented in this study was used for ECG signal classification, we made some intelligent modifications in its structure to achieve the goals of the current research. Beside the diversity of classifications and representation of an adaptive CNN, we studied the effect of baseline changes that caused 5% reduction in accuracy (on average) because of difference in mindsets and baselines of different subject. Changing the structure of the CNN parameters provided two main significant advantages: model improvement with higher subsampling rate and achieving the faster run-time. Making a shorter stride helped the model to reach acceptable accuracies in both training and testing.